\documentclass[acmsmall,nonacm]{acmart}

\AtBeginDocument{%
  }

\acmJournal{JACM}
\acmVolume{37}
\acmNumber{4}
\acmArticle{0}
\acmMonth{8}

\newcommand{\etal}{\textit{et al}.}
\newcommand{\ie}{\textit{i}.\textit{e}.}
\newcommand{\eg}{\textit{e}.\textit{g}.}

\newcommand{\wwj}[1]{{\color{black}#1}}
\newcommand{\linll}[1]{{\color{black}#1}}
\newcommand{\fzj}[1]{{\color{black}#1}}
\newcommand{\wj}[1]{{\color{black}#1}}

\newcommand{\rv}[1]{{\color{black}#1}}

\usepackage{graphicx}
\usepackage{subfigure}
\usepackage{hyperref}
\usepackage{balance}
\usepackage{booktabs}
\usepackage{color}
\usepackage{multirow}
\usepackage{xcolor}
\usepackage{caption}

\begin{document}

\title{Semi-Supervised Learning for Mars Imagery Classification and Segmentation}

\author{Wenjing Wang}
\email{daooshee@pku.edu.cn}
\author{Lilang Lin}
\email{linlilang@pku.edu.cn}
\author{Zejia Fan}
\email{zejia@pku.edu.cn}
\author{Jiaying Liu}
\email{liujiaying@pku.edu.cn}

\authornote{Corresponding Author. This work was supported by the National Key Research and Development Program of China under Grant No. 2018AAA0102702, and the National Natural Science Foundation of China under Contract No.62172020.}

\affiliation{%
  \institution{Wangxuan Institute of Computer Technology, Peking University}
  \streetaddress{No. 128 Zhongguancun North Street}
  \city{Haidian District}
  \state{Beijing}
  \country{China}
}

\renewcommand{\shortauthors}{Wang, et al.}

\begin{abstract}
\wwj{With the progress of Mars exploration, numerous Mars image data are collected and need to be analyzed.
\rv{However, due to the severe train-test gap and quality distortion of Martian data}, the performance of existing computer vision models is unsatisfactory.
In this paper, we \wj{introduce} a semi-supervised framework for \wj{machine vision on Mars} and try to resolve two specific tasks: classification and segmentation.
Contrastive learning is a powerful representation learning technique. \fzj{However, there is too much information overlap between Martian data samples, leading to a contradiction between contrastive learning and Martian data.}
Our key idea is to reconcile this contradiction with the help of annotations and further take advantage of unlabeled data to improve performance.
For classification, we propose to ignore inner-class pairs on labeled data as well as neglect negative pairs on unlabeled data, forming supervised inter-class contrastive learning and unsupervised similarity learning.
For segmentation, we extend supervised inter-class contrastive learning into an element-wise mode and use online pseudo labels for supervision on unlabeled areas.
Experimental results show that our learning strategies can improve the classification and segmentation models by a large margin and outperform state-of-the-art approaches.}
\end{abstract}



\maketitle

\section{Introduction}

\wwj{Humanity's interest in the universe is prevalent and enduring. 
In recent years, machine learning has shown its great power in space exploration.
For example, the first black hole image was captured by combining data from eight telescopes using a machine learning algorithm~\cite{blackhole}.
As techniques develop, machine learning will play a more and more significant role in scientific fields.} 

\wwj{Humans have been exploring the planet Mars since the last century.
Multiple rovers have been dispatched to Mars, sending an enormous amount of images to earth.
With these massive images, data-driven learning is increasingly being used in Mars research.
Wagstaff \etal~\cite{WagstaffLSGGP18} proposed to automatically classify images from the Mars Science Laboratory (MSL) mission with a neural network.
Targeting rover self-driving, Swan \etal~\cite{AI4Mars} explored the task of Mars terrain segmentation.
However, these works simply applied conventional machine learning algorithms designed for Earth object classification or street scene segmentation.
\rv{They neglect the properties of Mars data and leave the features of extraterrestrial planet surfaces unexplored.}
}

\fzj{Mars data provides specific difficulties for machine learning methods. Firstly, along with the exploration progress, the rover moves to the new area and collects new data, which can cause severe \textbf{train-test gap}. Secondly, the
quality of Martian data suffers in many ways, such as bad weather conditions, camera equipment damage, and signal loss in Mars-to-Earth transmission, which gives rise to \textbf{limited information quality}.}
\rv{Detailed analysis and visualization will be given in Section~\ref{sec:dataset}.}

\wwj{Many methods have been proposed to narrow train-test gaps.
Some works revise loss designs~\cite{tripletloss,focalloss,centerloss}, some focus on imbalanced data distribution~\cite{Decoupling,bbn,CuiJLSB19,CaoWGAM19}, some propose specific training strategies~\cite{Dropout,MixUp,CutMix}.
Although these approaches are powerful for common vision tasks, the train-test gap on Mars rover data is too challenging \rv{as we will show in Section~\ref{sec:dataset}}, making existing methods ineffective.}

\wwj{\wj{Data quality improvement} is a popular topic.
For image quality, researchers have \fzj{studied methods to deal with various kinds of distortion}, including but not limited to super-resolution~\cite{srcnn,jin2021multi}, de-noising~\cite{BM3D,yan2020depth}, and illumination enhancement~\cite{MSR}.
However, Mars rover images suffer from compound distortions, which are too complex for existing restoration methods to handle.
For data diversity, image augmentation~\cite{MixUp,CutMix} is not powerful enough, while adversarial-learning-based image generation~\cite{abs-1904-09135} is not reliable.
For annotation quality, researchers have explored how to train with noisy labels~\cite{LuFXHWG17} or synthesize more labels~\cite{Pseudo}, but these approaches are not effective enough on Mars rover data \rv{as we will show in Section~\ref{sec:exp_cls}}.}

\wwj{Unlike existing methods, we solve the problem through representation learning.
With a robust visual representation, the train-test gap and low data quality can be resolved simultaneously.
Based on this methodology, we study two \wj{Martian} vision tasks: image classification and semantic segmentation. The former is about image-level prediction, while the latter is about per-pixel prediction.
Classification and segmentation are very representative tasks.
Our exploration of them can also provide insights for other \wj{Martian} vision tasks, such as object detection, \fzj{tracking, and locating}.}

\wwj{We adopt a widely-used representation learning approach: contrastive learning~\cite{MoCo,chen2020simple}. 
Contrastive learning increases the mutual information between positive pairs and decreases the similarity between negative pairs. 
It can improve the separability and compactness of features, providing a more suitable representation space for various downstream vision tasks.
However, directly applying it to Mars rover data results in poor performance.
This is because there is a severe information overlap between different Mars data samples, which negates the effect of contrastive learning.}

\begin{figure}[t]
    \centering
    \centering
    \includegraphics[width=0.99\linewidth]{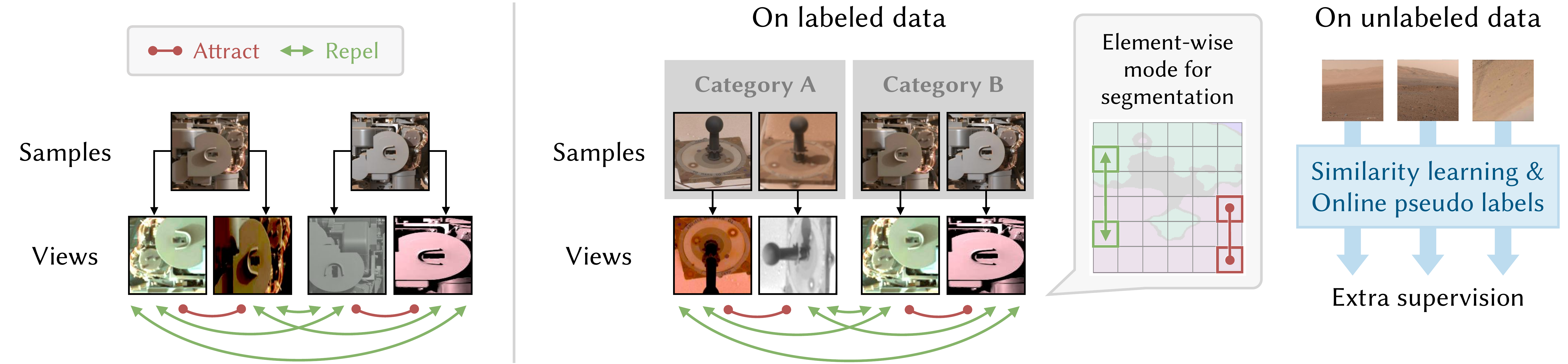}
    \caption{The proposed semi-supervised learning paradigm for \wj{Martian} machine vision \wj{tasks}. Left: the framework of conventional contrastive learning. Due to the redundancy in Mars data, naive pairs can be inappropriate.
    Right: our semi-supervised framework. On labeled data, we generate inner-class positive pairs and inter-class negative pairs. On unlabeled data, we use similarity learning \wj{and} online pseudo labels to introduce more supervision.}
    \label{fig:teaser}
\end{figure}

\wwj{To resolve this contradiction, we propose a semi-supervised learning strategy. On the one hand, we make use of annotations and ignore pairs within the same class, forming supervised inter-class contrastive learning. On the other hand, we train models on unlabeled images or areas to introduce more supervision.
More specifically, for Mars image classification, we abandon negative pairs and carry out unsupervised similarity learning on unlabeled images.
For segmentation, we further revise contrastive learning into a pixel-wise mode with online pseudo labels on unlabeled areas.
Experimental results demonstrate that our method achieves superior performance for Mars rover imagery classification and segmentation.}

\wwj{Our contributions can be summarized as follows:
\begin{itemize}
    \item \wj{Targeting at Martian machine vision tasks
}, we propose a semi-supervised learning framework, which outperforms existing approaches by a large margin in terms of classification and segmentation.
    \item For Mars imagery classification, we propose supervised inter-class contrastive learning and unsupervised similarity learning. By abandoning inter-class pairs on labeled data as well as negative pairs on unlabeled data, we resolve the contradiction between Mars rover images and contrastive learning.
    \item For Mars imagery segmentation, we extend inter-class contrastive learning into an element-wise mode and introduce online pseudo labels on the unlabeled area. Our method not only suits per-pixel prediction tasks but also makes use of the unlabeled area for further supervision, improving the performance of segmentation.
\end{itemize}}

\wwj{The rest of the article is organized as follows.
Section~\ref{sec:related} provides a detailed review of the relevant literature.
Section~\ref{sec:dataset} presents an in-depth analysis of Mars rover data.
Sections \ref{sec:classification} and \ref{sec:segmentation} introduce the proposed semi-supervised frameworks for classification and segmentation, respectively.
Experimental results and analyses are in Sections~\ref{sec:exp_cls} and \ref{sec:exp_seg}.
Concluding remarks are finally given in Section~\ref{sec:conclusion}.}

\section{Related Works}
\label{sec:related}

\subsection{Machine Learning in Mars Exploration}

\wwj{Machine learning has been utilized for a variety of planetary science tasks, such as exoplanet detection \cite{Shallue_2018}, comparative planetology and exoplanet biosignatures \cite{walker2018exoplanet}. Readers may refer to \cite{azari2020integrating} for a more comprehensive review and outlook.}


\wwj{For Mars exploration, existing machine learning applications can be categorized into two categories: in-situ (Mars edge) and ex-situ (Earth edge)~\cite{ML4Mars}.
For in-situ \wj{methods}, machine learning can benefit autonomous decision-making and save bandwidth by filtering out undesired images.
The Opportunistic Rover Science (OASIS) framework uses machine learning algorithms to identify terrain features~\cite{Mars_8,Mars_26}, dust devils, and clouds~\cite{Mars_30}.
For rover navigation, Abcouwer \etal~\cite{Mars_28} presented two heuristics to rank candidate paths, where a machine learning model is applied to predict untraversable areas.
For ex-situ \wj{methods}, machine learning can help scientists analyze data and notify noteworthy findings.
JPL scientists~\cite{Mars_31} built an impact crater classifier to analyze images captured by the Martian Reconnaissance Orbiter.
Dundar \etal~\cite{Mars_32} applied machine learning algorithms to discover less common minerals and search for aqueous mineral residue.
Rothrock \etal~\cite{Mars_33} designed machine learning models to identify terrain types and features in orbital and ground-based images.
The analysis can alert areas of slippage for rovers and assist the determination of potential landing sites for new missions.
Wagstaff \etal~\cite{WagstaffLSGGP18} created a dataset of the Mars surface environment and trained AlexNet~\cite{alexnet} for content classification.
Swan \etal~\cite{AI4Mars} collected a terrain segmentation dataset of $35K$ high-resolution images and tested the performance of DeepLabv3+~\cite{DeeplabV3_plus}.
\wj{To find an} efficient energy distribution between different systems in the orbiter, Petkovic \etal~\cite{Mars_37} applied multi-target regression to estimate the power consumption of the thermal system.}
\rv{For geomorphic mapping, Wilhelm~\cite{WilhelmGPSWWW20} built a dataset and provided an automated landform analysis strategy.}

However, most of the aforementioned algorithms are non-deep, taking no advantage of powerful neural networks.
Some research builds neural networks but directly applies models designed for conventional computer vision tasks~\cite{WagstaffLSGGP18,AI4Mars}, thus having unsatisfactory performance.
\rv{Moreover, advanced learning strategies such as semi-supervised and weakly supervised learning in the Martian scenario remains unexplored. The ``weak supervision'' in \cite{WilhelmGPSWWW20} refers to window sliding with Markov Random field smoothing for creating maps, which is far different from learning representation with limited data.}
In \wj{this} paper, we present an in-depth analysis of images captured by Mars rovers and \wj{introduce} a more powerful semi-supervised framework, \rv{which expands the research of deep learning for Mars.}


\subsection{Improving Classification and Segmentation Performance}

\wwj{Classification and segmentation are some of the most basic and popular tasks in computer vision. There have been many techniques for improving their performance.}

\wwj{Several works propose loss designs to balance positive and negative samples.}
Triplet loss~\cite{tripletloss} minimizes the distance between positive pairs and maximizes the distance between negative pairs.
Center loss~\cite{centerloss} clusters the feature representation.
Focal Loss~\cite{focalloss} aims at the imbalance between positive and negative samples.
Classification problems often suffer from data imbalance across classes.
\fzj{To solve data imbalance, re-sampling and re-weighting-based methods~\cite{Decoupling,CuiJLSB19,CaoWGAM19} are proposed.
In work \cite{Decoupling}, researchers decouple the learning procedure into representation learning and classification, then apply class-balanced sampling for classifier retraining. CB Loss~\cite{CuiJLSB19} represents the additional benefit with a hyper-parameter related to the sample volume and re-weighting the samples based on the additional benefit. LDAM~\cite{CaoWGAM19} aims to minimize the margin-based generalization bound along with prior re-weighting or re-sampling strategy.}
CutOut~\cite{devries2017improved}, CutMix~\cite{CutMix}, and ClassMix~\cite{olsson2021classmix} are powerful data augmentation mechanisms.
\wwj{CutOut~\cite{devries2017improved} randomly masks out regions, removing contiguous sections of images.
The limitation is that CutOut only captures relationships within the samples.
To introduce information across samples, CutMix~\cite{CutMix} combines different parts of two images to generate a new image.}
\linll{However, the random combination may destroy the semantic structure of the original image.
Thus, ClassMix~\cite{olsson2021classmix} combines semantic classes extracted from different images to make the generated image more meaningful.}
ReCo~\cite{liu2021bootstrapping} is a contrastive learning framework designed at a regional level to assist learning in semantic segmentation. \linll{It is computationally expensive to carry out pixel-level contrastive learning for all available pixels in high-resolution training.} \wwj{To reduce memory requirements,} \linll{ReCo introduces an active hard sampling strategy to optimize only a few queries and keys.}

\rv{However, all these methods have limited effectiveness on Mars rover data as we will demonstrate in Sections~\ref{sec:exp_cls} and \ref{sec:exp_seg}.}
In this paper, we propose a more effective representation learning strategy and achieve superior performance.

\section{Mars Imagery Datasets}
\label{sec:dataset}

In this paper, we experiment with our semi-supervised framework for on two specific Martian vision tasks: image classification and segmentation.

As for classification, we apply the MSL surface dataset~\cite{WagstaffLSGGP18}.
Wagstaff \etal~\wwj{collected} 6691 images by three instruments of the Curiosity Rover.
The Mars rover mission scientist defined 24 categories for the dataset.
Training and evaluation are split by Mars solar day (sol).
Data on sol 3-564 is for training and validation, while sol 565–1060  for testing.
Different from \cite{WagstaffLSGGP18}, we reshuffle the training and validation sets \rv{ to narrow the train-val gap, which can improve the top-1 accuracy by about 2\%.}
Our testing set remains the same \cite{WagstaffLSGGP18}.

As for segmentation, we apply the AI4Mars dataset~\cite{AI4Mars}, a large-scale Mars dataset for terrain classification and segmentation. This dataset consists \wwj{of} $35K$ images from Curiosity, Opportunity, and Spirit Rovers, collected through crowdsourcing. \wwj{Each label references the views of approximately ten people to ensure the annotation quality.}
Considering that the images obtained during the actual Mars exploration must be associated with the rovers' progress, we reasonably rearrange the dataset in the chronological order taken, just as the setting followed by the classification dataset. In ascending order of the shooting date, images in the training and validation sets account for the first 60\% of the dataset, \wj{\ie,} sol 1-1486, then the test data for the last 40\%, \wj{\ie,} sol 1487-2579.

\begin{figure}[t]
    \centering
    \centering
    \includegraphics[width=0.99\linewidth]{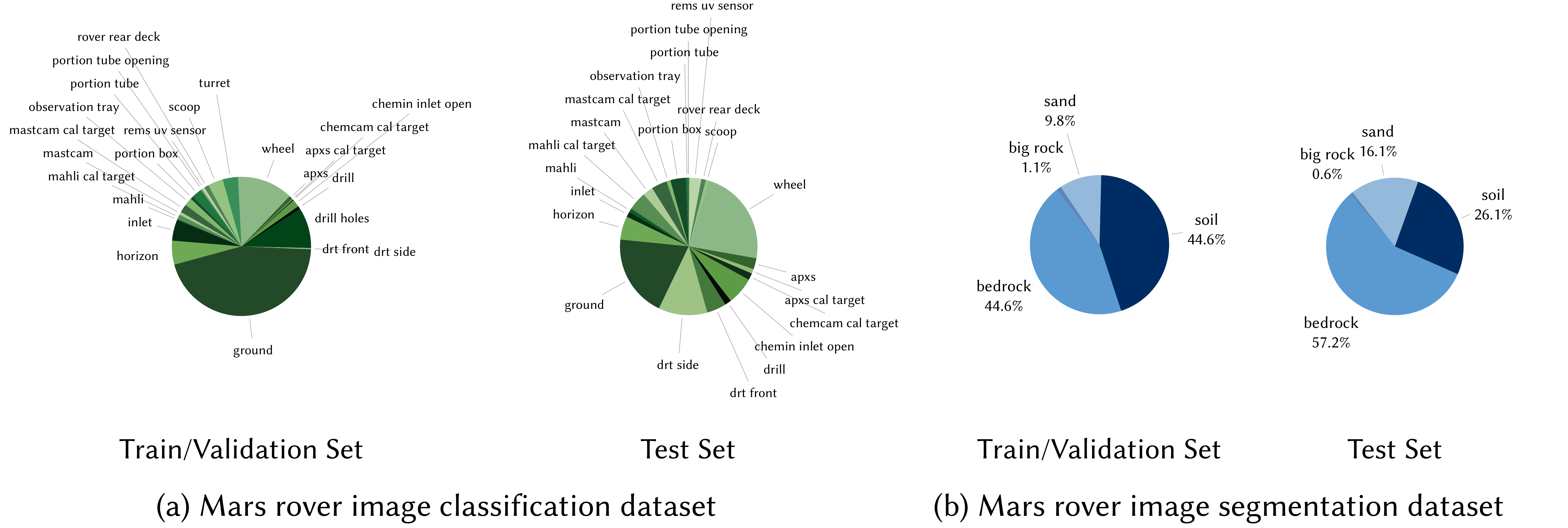}
    \caption{The uneven category distribution of train/validation and test sets in Mars rover data. (a) MSL surface dataset~\cite{WagstaffLSGGP18}. (b) AI4Mars dataset~\cite{AI4Mars}.}
    \label{fig:gap_distribution}
\end{figure}

\begin{table}[t]
  \centering
  \caption{The number of samples for each category in MSL surface dataset~\cite{WagstaffLSGGP18}.}
  \label{table:dataset_cls}
\begin{minipage}{0.4\textwidth}
    \begin{tabular}{l|c|c}
    \toprule
     & Train/Val & Test\\
     \midrule
    apxs & 46 & 34\\
    apxs cal target & 10 & 14\\
    chemcam cal target & 15 & 21\\
    chemin inlet open & 92 & 84\\
    drill & 39 & 20\\
    drill holes & 506 & 0\\
    drt front & 6 & 60\\
    drt side & 12 & 150\\
    ground & 2430 & 254\\
    horizon & 299 & 72\\
    inlet & 261 & 16\\
    mahli & 14 & 12\\
    \bottomrule
    \end{tabular}
\end{minipage}
\begin{minipage}{0.4\textwidth}
    \begin{tabular}{l|c|c}
        \toprule
        & Train/Val & Test\\
        \midrule
        mahli cal target & 60 & 57\\
        mastcam & 36 & 32\\
        mastcam cal target & 105 & 48\\
        observation tray & 99 & 12\\
        portion box & 38 & 48\\
        portion tube & 128 & 9\\
        portion tube opening & 20 & 2\\
        rems uv sensor & 32 & 36\\
        rover rear deck & 57 & 14\\
        scoop & 190 & 10\\
        turret & 193 & 0\\
        wheel & 698 & 300\\
        \bottomrule
    \end{tabular}
\end{minipage}
\end{table}

\begin{table}[t]
  \centering
  \caption{The total area for each category in AI4Mars dataset~\cite{AI4Mars}.}
  \label{table:dataset_seg}
    \begin{tabular}{l|c|c|c|c}
    \toprule
    & soil & bedrock & sand & big rock \\
    \midrule
    Train/Val   & $2.59 \times 10^9$    & $2.59\times 10^9$ & $5.68\times 10^8$ & $6.35\times 10^7$ \\
    Test        & $9.57 \times 10^8$   & $2.10\times 10^9$ & $5.89\times 10^8$ & $2.28\times 10^7$ \\
    \bottomrule
    \end{tabular}
\end{table}

\begin{figure}[t]
    \centering
    \centering
    \includegraphics[width=0.99\linewidth]{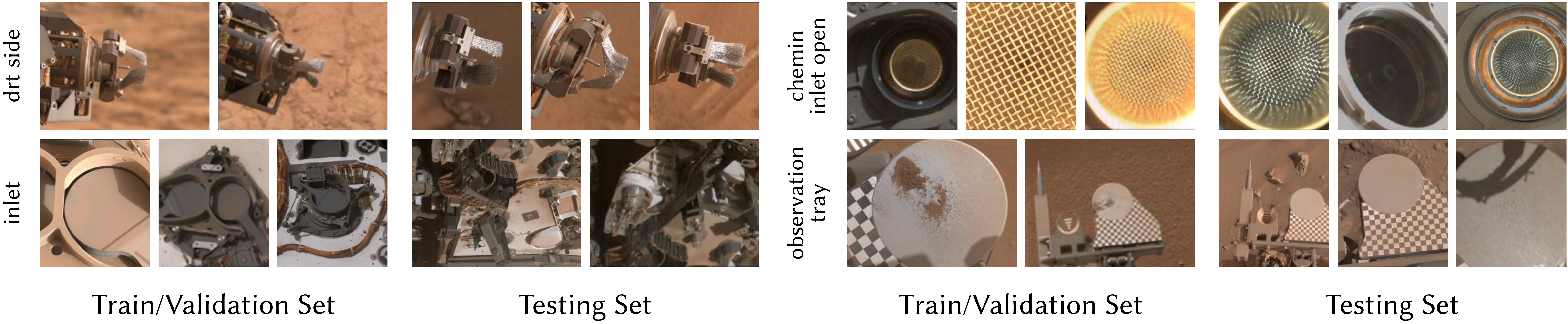}
    \caption{The various appearance of the same objects in the MSL surface dataset~\cite{WagstaffLSGGP18}.}
    \label{fig:gap_appearance}
\end{figure}

\begin{figure}[t]
    \centering
    \centering
    \includegraphics[width=0.99\linewidth]{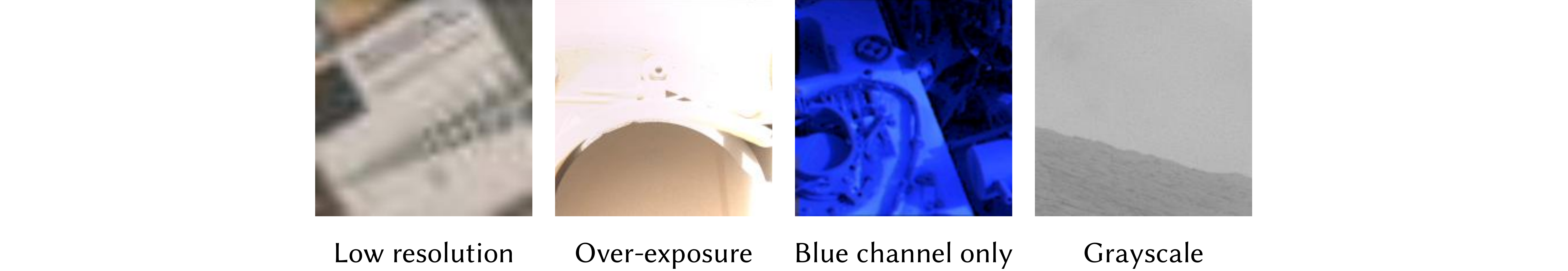}
    \caption{Examples of low-quality images in the MSL surface dataset~\cite{WagstaffLSGGP18}.}
    \label{fig:quality_image}
\end{figure}

\begin{figure}[t]
    \centering
    \centering
    \includegraphics[width=0.99\linewidth]{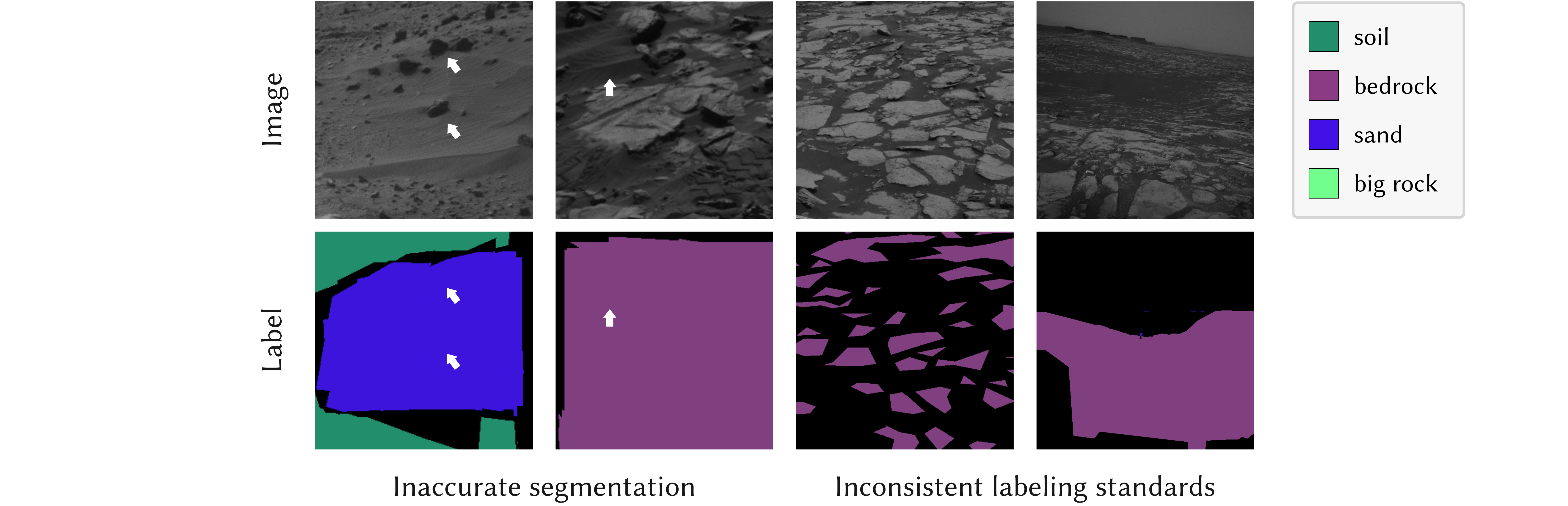}
    \caption{Some annotations are of low quality in the AI4Mars dataset~\cite{AI4Mars}. On the left, as indicated by white arrows, rocks are labeled as sand (the first column), while a patch of sand is labeled as bedrock (the second column). On the right, similar terrains are labeled inconsistently on different samples.}
    \label{fig:quality_label}
\end{figure}

Generally, Mars rover data poses two challenges for machine learning: 

\begin{itemize}
    \item \textbf{Train-Test Gap}.
    In actual missions, the training and validation processes can only use data collected in the past, while future data is the testing target.
    However, since rovers capture images at a non-uniform frequency and keep traveling to new areas, the collected data varies over time.
    \rv{This feature leads to a large train-test gap for class distribution and object appearance.}
    As shown in Fig.~\ref{fig:gap_distribution}~(a) and Table~\ref{table:dataset_cls}, the classification dataset suffers from unbalanced class distribution. The samples of \textit{drt side} class are largely absent from the training set, though appear frequently in the testing set. On the contrary, classes like \textit{turrent}, \textit{scoop} and \textit{drill holes} have abundant samples in the training set but scarcely appear in the testing set. The same thing happens on the segmentation dataset,
    As in Fig.~\ref{fig:gap_distribution}~(b) and Table~\ref{table:dataset_seg}, while the proportion of the \textit{sand} class increases markedly, \wwj{the proportion of} the \textit{soil} class decreases.
    \wwj{The gap lies in not only class distribution but also object appearance} as shown in Fig.~\ref{fig:gap_appearance}, \eg, the samples of \textit{drt side} class in the training set were shot from a distance, while those in the testing set were shot up close. The case for the \textit{inlet} class is the opposite. 

    \item \textbf{Limited Information Quality}.
    The quality of Martian data suffers in many ways.
    First, rover data may be affected by wrong shooting operations, camera equipment damage, and signal loss in Mars-to-Earth transmission.
    These errors can degrade the visual quality of images, as shown in Fig.~\ref{fig:quality_image}.
    Second, because of the monotonous Mars scenes and limited data sources, Mars datasets are usually less diverse than common computer vision datasets.
    Third, annotating Mars data requires particular expert knowledge. High labeling costs and limited budgets hamper the acquisition of high-quality annotation. Accordingly, the label quality of Mars rover data can be unsatisfactory as shown in Fig.~\ref{fig:quality_label}.
\end{itemize}

\section{Semi-supervised Mars Imagery Classification}
\label{sec:classification}

\wwj{We first introduce our solution for the Mars imagery classification task.
To make full use of annotations and unlabeled images, we design a semi-supervised contrastive learning scheme, consisting of two sub-strategies: supervised inter-class contrastive learning and unsupervised similarity learning.
In the following, we first review contrastive learning, then introduce our specific method.}

\subsection{Review of Contrastive Learning}


\wwj{As stated in \cite{chen2020simple}, the core of contrastive learning is to increase the mutual information between pairs of positive samples and decrease the similarity between pairs of negative samples.
To generate positive and negative samples, data transformation is widely utilized. 
Specifically, for a sample $\mathbf{x}_i$ from the dataset $\mathbf{X}$, contrastive learning performs random data transformation $\mathcal{T}$ to obtain transformed data $\mathbf{x}^1_i=\mathcal{T}^1(\mathbf{x}_i)$, $\mathbf{x}^2_i=\mathcal{T}^2(\mathbf{x}_i)$, where $\mathbf{x}^1_i \neq \mathbf{x}^2_i$ are two different views of $\mathbf{x}_i$, and $\mathcal{T}^1$, $\mathcal{T}^2$ are transformations sampled from $\mathcal{T}$ independently.
With $\mathcal{T}$, contrastive learning assigns positive pairs as views of the same image and negative pairs as views of different images.
Then, a feature encoder $f$ is used to extract the representation of $\mathbf{x}^1_i$ and $\mathbf{x}^2_i$, denoted as $\mathbf{z}^1_i = f(\mathbf{x}^1_i)$, $\mathbf{z}^2_i = f(\mathbf{x}^2_i)$.
Finally, contrastive learning normalizes the features into a spherical manifold and computes the cosine similarity between positive pairs and negative pairs.
The InfoNCE loss~\cite{infonce} is applied for optimization:
\begin{align}
    & \mathcal{L} = - \mathbb{E}_{\mathbf{x}_i} \left[ \log \frac {\exp(K(\mathbf{z}^1_i, \mathbf{z}^2_i) / \tau)} {\sum_{\mathbf{x}_j \in \mathbf{X}} \exp(K(\mathbf{z}^1_i, \mathbf{z}^2_j) / \tau)} \right],
    \label{eq:contrastive} \\
    & K(\mathbf{u},\mathbf{v}) = \mathbf{u}^\text{T} \mathbf{v} / |\!|\mathbf{u}|\!||\!|\mathbf{v}|\!|, \nonumber
\end{align}
where $\tau$ is a temperature hyper-parameter, and $K(\cdot,\cdot)$ is the kernel function for computing cosine similarity.}

\wwj{Although contrastive learning is originally designed for unsupervised learning, it can also assist supervised learning. Compared with a solo classification loss, training with an extra contrastive loss can enrich the visual representation, improving the robustness of the neural network.}

\begin{figure}[t]
    \centering
    \includegraphics[width=0.99\linewidth]{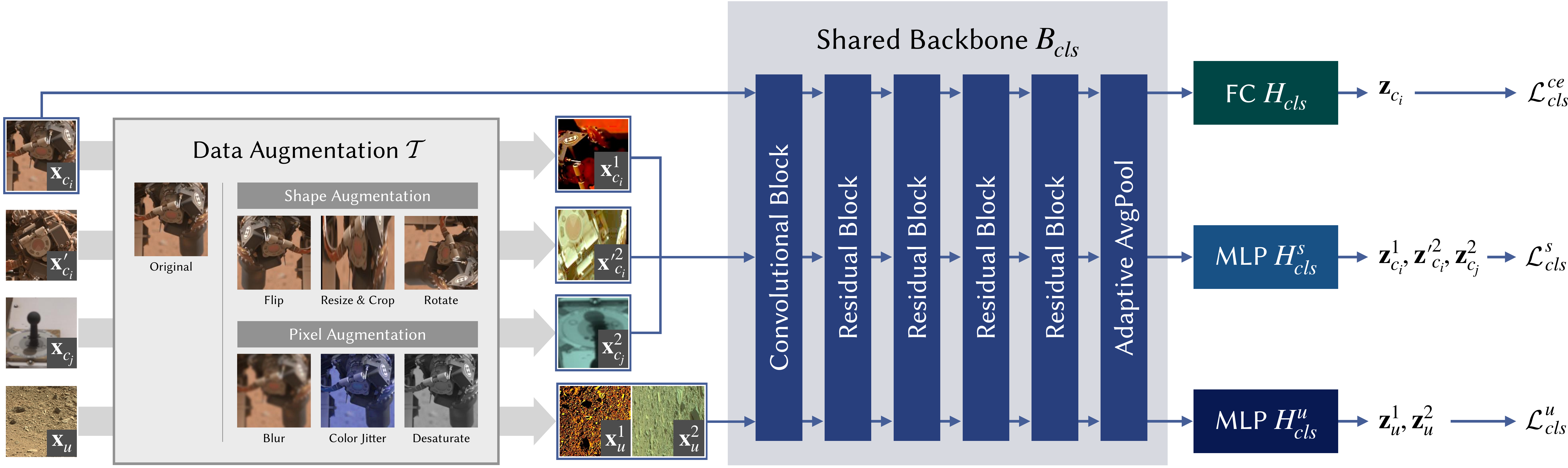}
    \caption{\wwj{The framework of our semi-supervised Mars image classification. There are three streams in the framework: (top-to-bottom) classification, supervised inter-class contrastive learning, and unsupervised similarity learning. Each stream shares the backbone and has independent heads.}}
    \label{fig:classification}
\end{figure}

\subsection{Supervised Inter-class Contrastive Learning}

\wwj{Current research mostly applies contrastive learning on large-scale Earth image datasets such as ImageNet~\cite{ImageNet} and JFT~\cite{JFT}, where the difference between images is large enough.
However, compared with diverse Earth scenes, Mars scenes are rather homogeneous.
Moreover, Mars rovers may take photos at the same scene multiple times, \eg, when investigating the surrounding terrain or monitoring equipment degradation.
These factors result in a severe information overlap between different Mars image samples. Since contrastive learning relies on the mutual information between different samples, cross-image information overlap may lead to the futility of contrastive learning on Mars data.}

\wwj{To address this problem, we make use of classification annotations to select more appropriate contrastive pairs.
Specifically, we delete negative samples belonging to the same class and add positive pairs of different samples belonging to the same class, \ie, we increase the mutual information of all samples in the same category and decrease the similarity only for pairs of different categories.
In other words, we turn the original unsupervised sample-wise contrastive learning strategy into a supervised inter-class version.}

\wwj{Denoting the number of categories as $C$, our supervised inter-class contrastive loss is:
\begin{equation}
    \mathcal{L}^s_{cls} = - \mathbb{E}_{\mathbf{x}_{c_i}} \left[ \log \frac {\exp(K(\mathbf{z}^1_{c_i}, \mathbf{z}'^{2}_{c_i}) / \tau)} {\sum_{c_j=1}^C \exp(K(\mathbf{z}^1_{c_i}, \mathbf{z}^2_{c_j}) / \tau)} \right],
\end{equation}
where \wj{$\mathbf{z}^1_{c_i}=f(\mathcal{T}^1(\mathbf{x}_{c_i}))$ and $\mathbf{z}'^2_{c_i}=f(\mathcal{T}^2(\mathbf{x}'_{c_i}))$}. $\mathbf{x}_{c_i}$ and $\mathbf{x}'_{c_i}$ are samples belonging to category $c_i$.}

\wwj{The selection of data augmentation $\mathcal{T}$ is one of the keys in contrastive learning.
The augmentation we use can be categorized into two types: shape and pixel.
Shape augmentation teaches the model to perceive objects under different camera angles and magnifications.
It contains random flipping, cropping, resizing, and rotation.
Pixel augmentation aims to improve the model's robustness to image quality degradations.
It contains Gaussian blur, color jittering, and desaturation.
With revised contrastive learning and targeted augmentation, the problems of data imbalance and low image quality can be greatly alleviated.}


\subsection{Unsupervised Similarity Learning}

\wwj{Our supervised inter-class contrastive loss relies on sufficient annotations.
However, labeling data requires a lot of manpower and financial resources.
For Mars data, the cost is particularly high.
Identifying Martian landscapes and rover components requires expert knowledge~\cite{WagstaffLSGGP18}.
Although annotations are expensive, pure images are relatively easy to obtain. 
Through past and current missions, scientists have acquired millions of images from Mars.
To reduce the reliance on annotations, we explore how to use unlabeled data to further improve classification performance.}

\wwj{Unlabeled Mars data also has information overlap between different samples. When applying contrastive learning to these data, negative pairs may come from the same scene and should not be forced apart in the feature space.
The only thing guaranteed is that different views of the same image should have similar representations.
Therefore, we adopt similarity learning, where we abandon negative pairs and only consider positive ones.
Denoting $\mathbf{x}_u$ as a sample from unlabeled data, the proposed unsupervised similarity loss is:
\begin{equation}
    \mathcal{L}^u_{cls} = - \mathbb{E}_{\mathbf{x}_u} \left[ K(\mathbf{z}^1_u, \mathbf{z}^2_u) \right].
\end{equation}

Why similarity learning does not lead to model collapse is an interesting question.
In experiments, we find that training classification with $\mathcal{L}_{cls}^{u}$ does cause collapse.
However, when we add $\mathcal{L}^s_{cls}$, model collapse is prevented.
This may be because supervised inter-class contrastive learning provides a good restriction to the feature representation, counteracting the bad impacts of similarly learning.}

\subsection{Full Model}

\wwj{As shown in Fig.~\ref{fig:classification}, our framework consists of three streams: classification, supervised inter-class contrastive learning, and unsupervised similarity learning.
Since we combine supervised and unsupervised learning, our approach is ``semi-supervised''.}

\wwj{The classification objective $\mathcal{L}^{ce}_{cls}$ is cross entropy.
Given \rv{a labeled} sample $\mathbf{x}_{c_i} \in \mathbf{X}$ of category $c_i$, $\mathcal{L}^{ce}_{cls}$ is:
\begin{equation}
    \mathcal{L}^{ce}_{cls} = - \mathbb{E}_{\mathbf{x}_{c_i}} \left[ \log \frac {\exp(\mathbf{z}_{c_i, c_i})} { \sum_{c_j=1}^C \exp(\mathbf{z}_{c_i, c_j})} \right],
\end{equation}
where $\mathbf{z}_{c_i, c_j}$ is the $c_j$-th element of $\mathbf{z}_{c_i}$, representing the prediction of the sample belonging to the $c_j$ category.}

\wwj{Our full loss function is:
\begin{equation}
    \mathcal{L}_{cls} = \mathcal{L}^{ce}_{cls} + \lambda^s_{cls} \mathcal{L}^s_{cls} + \lambda^u_{cls} \mathcal{L}^u_{cls},
\end{equation}
where $\lambda^s_{cls}$ = $1$ and $\lambda^u_{cls}$ = $0.2$ are hyper-parameters to balance different training objectives. The temperature hyper-parameter $\tau$ in $\mathcal{L}^s_{cls}$ is set to $0.2$.}

\wwj{We use ResNet-50~\cite{resnet} for classification.
In $\mathcal{L}^s_{cls}$ and $\mathcal{L}^u_{cls}$, the feature encoder $f$ consists of a shared ResNet-50 backbone $B$ and a 2-layer Multi-Layer Perceptrons (MLP) head. We denote the MLP in $\mathcal{L}^s_{cls}$ and $\mathcal{L}^u_{cls}$ as $H^s_{cls}$ and $H^u_{cls}$, respectively. The output dimension of $H^s_{cls}$ and $H^u_{cls}$ is $128$.}

\section{Semi-supervised Mars Imagery Segmentation}
\label{sec:segmentation}

\wwj{In this section, we extend our semi-supervised contrastive learning scheme from Mars imagery classification to semantic segmentation.
The challenge is that classification is an image-wise prediction task while segmentation is pixel-wise.
In segmentation, each pixel has its own category, which requires the model to perceive objects at a finer scale.
However, conventional contrastive learning methods treat the input image as a whole.
To address this problem, we propose element-wise contrastive learning.}

\begin{figure}[t]
    \centering
    \includegraphics[width=0.99\linewidth]{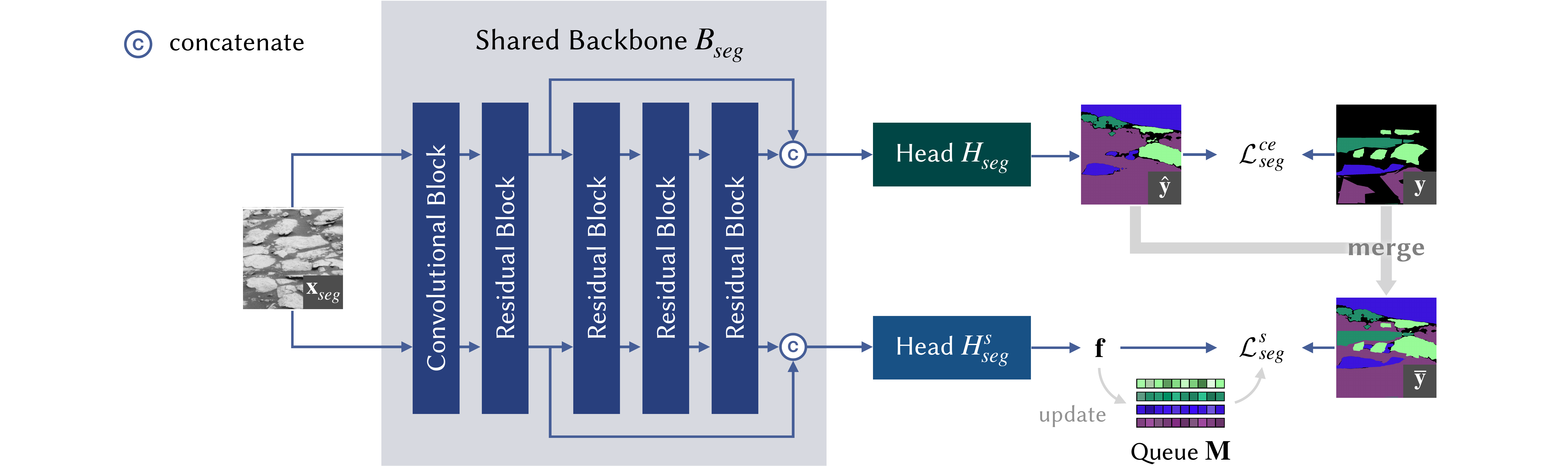}
    \caption{\wwj{The framework of our semi-supervised Mars image segmentation. There are two streams: (top-to-bottom) segmentation and element-wise inter-class contrastive learning. The two streams share the backbone and have independent heads.}}
    \label{fig:segmentation}
\end{figure}

\subsection{Element-wise Inter-class Contrastive Learning}

\wwj{Given an input image $\mathbf{x}_{seg}$, we first use a feature encoder $g$ to extract the representation $\mathbf{f} = g(\mathbf{x}_{seg})$.
In classification, the extracted representation is a vector, while in segmentation, what we obtain is a 2D feature map.
The resolution of $\mathbf{f}$ is usually lower than that of the input image $\mathbf{x}_{seg}$.
Then, we down-sample the segmentation annotation of $\mathbf{x}_{seg}$ to match the resolution of $\mathbf{f}$.
In this way, each spatial element in $\mathbf{f}$ can have its own category, which supports us to conduct element-wise inter-class contrastive learning.}

\wwj{Similar to our approach for classification, we want to
\linll{make features of the same category closer and features of different categories more separable.}
Denoting $\mathbf{f}_i$ as an element in the 2D feature map $\mathbf{f}$ and its category as $c_i$, we can have a na\"ive form of element-wise inter-class contrastive learning:
\begin{equation}
    \tilde{\mathcal{L}}^s_{seg} = - \mathbb{E}_{\mathbf{x}_{seg}} \mathbb{E}_{\mathbf{f}_i} \left[ \log \frac { \sum_{c_j = c_i} \exp(K(\mathbf{f}_i, \mathbf{f}_j) / \tau)} {\sum_j \exp(K(\mathbf{f}_i, \mathbf{f}_j) / \tau)} \right].
\end{equation}
For pixels without labels, we simply ignore them.}

\wwj{The problem is that $\tilde{\mathcal{L}}^s_{seg}$ is impossible to calculate.
If the resolution of $\mathbf{f}$ is $128\times128$, there will be 16,384 elements in $\mathbf{f}$. Computing the similarity among 16,384 vectors, \ie, 268 million vector multiplications, exceeds the capacity of current computation devices.
Moreover, $\tilde{\mathcal{L}}^s_{seg}$ only considers elements in a single image.
To make training feasible and introduce cross-image contrastive pairs, we use a memory bank to store the history average representation for each category.}

\wwj{The proposed memory bank $\mathbf{M}=\{\mathbf{m}_{c_i}\}_{c_i=1}^C$ consists of $C$ queues, where $C$ is the number of categories.
Each queue $\mathbf{m}_{c_i}$ constantly removes the oldest element and stores the average of all features labeled $c_i$ in the current $\mathbf{f}$.
Denoting $\overline{\mathbf{m}}_{c_i}$ as the average of $\mathbf{m}_{c_i}$, our element-wise inter-class contrastive learning loss is:
\begin{equation}
    \mathcal{L}^s_{seg} = - \mathbb{E}_{\mathbf{x}_{seg}} \mathbb{E}_{\mathbf{f}_i} \left[\log \frac {\exp(K(\mathbf{f}_{i}, \overline{\mathbf{m}}_{c_i}) / \tau)} {\sum_{c_j=1}^{C} \exp(K(\mathbf{f}_{i}, \overline{\mathbf{m}}_{c_j}) / \tau)} \right].
\end{equation}
With an assistant memory bank, $\mathcal{L}^s_{seg}$ reduces the computation complexity and introduces cross-image information by storing previous features.}

\subsection{Online Pseudo Labeling for Semi-supervised Learning}

\wwj{Annotating segmentation is laborious and time-consuming.
Accordingly, 44.83\% of the area does not have labels in the AI4Mars dataset, which limits the effect of our supervised inter-class contrastive learning.
Moreover, as we stated in Section~\ref{sec:dataset}, the quality of annotation in AI4Mars is not satisfactory.
To exploit numerous unlabeled pixels and refine annotation, we expand supervision from labeled areas to more areas by semi-supervised learning.}

\wwj{To construct positive and negative pairs on unlabeled data, previous methods apply clustering~\cite{SwAV} or data transformation~\cite{MoCo}.
However, these strategies rely on sufficient and diverse data, which are not suitable for Mars scenarios.
We instead use a fairly good segmentation model and predict pseudo labels on unlabeled data.
With the help of segmentation prediction, we can construct more accurate contrastive pairs.}

\wwj{We first train a segmentation network on fully supervised data.
Then, we use it to predict a category for each unlabeled pixel.
To ensure the quality of pseudo labels, we remove predictions with low confidence. The confidence threshold is set to 0.9. 
Then, pseudo labels are added to join our \textit{Element-wise Inter-class Contrastive Learning}.
Formally, for an input image $\mathbf{x}_{seg}$ with ground truth annotation $\mathbf{y}$ and segmentation prediction $\hat{\mathbf{y}}$, we supplement the unlabeled part in $\mathbf{y}$ with the high-confidence part in $\hat{\mathbf{y}}$.
The merged result is denoted as $\overline{\mathbf{y}}$.
For fully supervised learning, $\mathcal{L}^s_{seg}$ refers $\mathbf{y}$ for the category of each pixel, while in semi-supervised learning, we replace $\mathbf{y}$ with $\overline{\mathbf{y}}$.
In this way, we introduce unlabeled data into training.
Semi-supervision can introduce more guidance to the network, making the feature space easier to generalize.}

\subsection{Full Model}

\wwj{Similar to classification, the segmentation objective $\mathcal{L}^{ce}_{seg}$ is also cross entropy:
\begin{equation}
    \mathcal{L}^{ce}_{seg} = - \mathbb{E}_{\mathbf{x}_{seg}} \mathbb{E}_{\mathbf{f}_i} \left[ \log \frac {\exp(\mathbf{f}_{i, c_i})} { \sum_{c_j=1}^{C} \exp(\mathbf{f}_{i, c_j})} \right],
\end{equation}
where $\mathbf{f}_{i, c_j}$ represents the prediction of this pixel belonging to the $c_j$ category, and $c_i$ is the ground truth label.}

\wwj{To make training stable and maximize the effectiveness of each learning design, we adopt a three-step training strategy: supervised contrastive learning pretraining, segmentation fine-tuning, and semi-supervised joint training.}

\wwj{We first pretrain the model with supervised contrastive learning alone, which provides a suitable feature space initialization for segmentation.
We apply \textit{Element-wise Inter-class Contrastive Learning} with ground truth annotations.}

\wwj{After pretraining, the model is trained with contrastive and segmentation losses simultaneously. The training objective is:
\begin{equation}
    \mathcal{L}_{seg} = \mathcal{L}^{ce}_{seg} + \lambda^s_{seg} \mathcal{L}^s_{seg},
\end{equation}
where $\lambda^s_{seg}$ controls the training balance between two objectives. To reduce the impact of contrastive learning, $\lambda^s_{seg}$ is set to $0.001$. 
The hyper-parameter $\tau$ in $\mathcal{L}^s_{seg}$ is set to $0.07$.}

\wwj{Finally, the model is trained with semi-supervised learning.
We add \textit{Online Pseudo Labeling} to element-wise inter-class contrastive learning.
The previous training steps ensure the accuracy of segmentation, providing good initial pseudo labels for this step.
With the training of the framework, we get better and better label estimates for unlabeled data, which promotes the semi-supervised learning process and enables the model to extract more separable feature representations.}

\wwj{The framework is shown in Fig.~\ref{fig:segmentation}.
Our model is based on DeepLabv3+~\cite{DeeplabV3_plus}.
The segmentation and contrastive streams share a same ResNet-101~\cite{resnet} backbone $B_{seg}$.
The heads in these two streams, $H_{seg}$ and $H^s_{seg}$, are all DeepLabv3+ segmentation heads.
The output dimension of $H^s_{seg}$ is 128.
The queue length in the memory bank $\mathbf{M}$ is 32.}




\section{Experiments for Mars Imagery Classification}
\label{sec:exp_cls}

\wwj{In this section, we evaluate our semi-supervised learning framework for classification.}
\wj{We first introduce the experimental settings, then show the comparison results against the state-of-the-arts, and finally provide ablation studies and more performance analysis.}

\subsection{Experiment Setup}

\wwj{For model training and evaluation, we use the MSL surface dataset~\cite{WagstaffLSGGP18}. 
\rv{For unsupervised similarity learning, we additionally collect 34k unlabeled color images from the NASA’s Planetary Data System (PDS)\footnote{https://pdsimage2.wr.usgs.gov/archive/MSL/}}.}

\wwj{We first pretrain the model on ImageNet~\cite{ImageNet} following MoCo V2~\cite{MoCoV2}.
Then, the model is fine-tuned on Mars data with Adam~\cite{Adam} optimizer for 30 epochs.
The mini-batch size is set to 16 for $\mathcal{L}^{ce}_{cls}$ and $\mathcal{L}^u_{cls}$ while 24 for $\mathcal{L}^s_{cls}$. 
Here, 24 equals the number of categories in the MSL surface dataset.
The initial learning rate is set to 1e-3 for $H_{cls}$ and 1e-6 for $B_{cls}$, $H^s_{cls}$ and $H^u_{cls}$, then multiplied by 0.1 at 20 and 25 epochs.
The fine-tuning process takes about 1 hour with an Nvidia GeForce RTX 2080Ti.}

\begin{table}[t]
  \centering
  \caption{Results for MSL rover image classification.}
  \vspace{1mm}
  \label{table:comparison}
    \begin{tabular}{l|l|c}
        \toprule
        Category & Method & Top-1 (\%)\\
        \midrule
        
        \multirow{2}{*}{Baseline} & AlexNet~\cite{alexnet} in \cite{WagstaffLSGGP18} & 66.70 \\
        & ResNet-50~\cite{resnet} &  79.28 $\pm$ 1.76\\
        
        \midrule
        
        & Focal loss~\cite{focalloss} & 82.86 $\pm$ 0.74 \\
        Loss design & Center loss~\cite{centerloss} & 82.91 $\pm$ 0.93 \\
        & Triplet loss~\cite{tripletloss} & 84.87 $\pm$ 1.13 \\
        
        \midrule
        
        \multirow{2}{*}{Training} & \rv{MixUp~\cite{MixUp}} & \rv{76.19 $\pm$ 1.65} \\
        \multirow{2}{*}{design} & \rv{CutMix~\cite{CutMix}} & \rv{80.61 $\pm$ 1.51} \\
        & \rv{Dropout~\cite{Dropout}} & \rv{83.37 $\pm$ 0.41} \\
        
        \midrule
        
        & S4L~\cite{s4l}, rotation & 75.19 $\pm$ 1.73 \\
        Semi- 
        & \rv{SupCon~\cite{SupCon} + Linear Classifier} & \rv{77.11 $\pm$ 2.89} \\
        supervised
        & Pseudo labeling~\cite{Pseudo} & 78.64 $\pm$ 0.04 \\
        learning
        & S4L~\cite{s4l}, jigsaw & 81.81 $\pm$ 2.33 \\
        & \rv{SupCon~\cite{SupCon} + All Layers} & \rv{83.22 $\pm$ 1.33} \\
        & \rv{SsCL~\cite{zhang2022semi}} & \rv{90.81 $\pm$ 1.85}\\
        
        \midrule
        
        \multirow{2}{*}{Re-sampling} & Decoupling~\cite{Decoupling}, cRT & 80.94 $\pm$ 1.43 \\
        & Decoupling~\cite{Decoupling}, LWS & 81.30 $\pm$ 0.62 \\
        
        \midrule
        
        \multirow{2}{*}{Re-weighting} & Class-balanced loss~\cite{CuiJLSB19} & 80.02 $\pm$ 0.89 \\
        & LDAM-DRW~\cite{CaoWGAM19} & 82.12 $\pm$ 1.92 \\
        
        \midrule
        
        \textbf{Ours} & & \textbf{95.86} $\pm$ 1.63 \\
          \bottomrule
    \end{tabular}
\end{table}

\subsection{Comparison Results}

We compare the performance of our model with the other ten related methods.
For reliability, we run each experiment three times, then show the mean and standard deviation in Table~\ref{table:comparison}.

We first carry out a comparison with work \cite{WagstaffLSGGP18}, in which the model performance is 66.70\%. The low performance may be due to the limited capability of AlexNet~\cite{alexnet}.
The performance improves to 79.28\% after we change the baseline to ResNet-50~\cite{resnet}, demonstrating the necessity of using good feature extractors.

As for the next step, we take works that aim to balance positive and negative samples through loss \rv{and training} designs into account. We \rv{first} consider three widely-used losses: Triplet loss~\cite{tripletloss}, Center loss~\cite{centerloss}, and Focal loss~\cite{focalloss}. 
\rv{These loss designs enforce the embedded distances among different categories and can improve the classification performance by 3$\sim$6\%.
However, they do not enrich the feature representation, therefore have limited effectiveness.}
\rv{Next, we consider three widely-used training strategies: MixUp~\cite{MixUp}, CutMix~\cite{CutMix}, and Dropout\cite{Dropout}. Their performances are below 85\%, indicating that data augmentation and model regularization cannot solve the problem of Martian classification.}

Since we introduce unlabeled data \fzj{into} our framework, the comparison also involves \rv{four} semi-supervised learning methods.
One is pseudo learning~\cite{Pseudo}, which generates pseudo labels on unlabeled data.
However, the performance degrades with pseudo learning, which is possible because the unlabeled data crawled from PDS is uncurated.
Compared with the MSL surface dataset, the collected unlabeled data can be more long-tailed and unbalanced.
Therefore, the pseudo labels on unlabeled data can mislead the classification model training to some extent.
\rv{The second one} is S4L~\cite{s4l}, which applies self-supervised learning on unlabeled data.
Among self-supervised tasks, the jigsaw pretext task improves the performance a little, while rotation brings down the classification accuracy.
This could be because most unlabeled data contains Martian soil and rock, which does not offer much semantic information. In that case, unlabeled data is ambiguous for the jigsaw and rotation pretext tasks.
\rv{The third one is SupCon~\cite{SupCon}. We test two training policies: unfreezing a linear classifier with a frozen model base (originally used in \cite{SupCon}) and jointly fine-tuning all layers. Although SupCon also restricts positive samples to be of the same class, our Supervised Inter-class Contrastive Learning is quite different from SupCon. First, SupCon simply selects training samples by random, while we control a training batch to equally cover all classes, which can efficiently cluster representation for classification. Second, SupCon splits the process of representation learning and classification. In comparison, our Supervised Inter-class Contrastive Learning is trained along with classification, which is more convenient to implement and better combines supervision information with contrastive learning. Because SupCon randomly selects training samples and splits the learning of representation and classification, it only achieves a top-1 accuracy of 77.11\% for unfreezing a linear classifier and 83.22\% for fine-tuning all layers, which is much lower than our 95.86\%.}
\rv{The fourth one is SsCL~\cite{zhang2022semi}, which is cross entropy and pseudo-label based contrastive learning with pseudo label propagation according to similarity.
However, the contrastive learning in SsCL is still unsupervised.
Also, the unlabeled data is utilized by pseudo label co-calibration with similarity alignment, which is in direct proportion to the label number of each class and thus cannot handle the long-tailed and unbalanced unlabeled Martian data.
Due to the unimproved contrastive learning scheme and the unsuitable unsupervised learning strategy, the top-1 accuracy of SsCL is only 90.81\%, which is >5\% worse than ours.
}

We also compare our methods with three state-of-the-art techniques for imbalanced data. Decoupling~\cite{Decoupling} is based on re-sampling, while Class-balanced loss~\cite{CuiJLSB19} and LDAM-DRW~\cite{CaoWGAM19} on re-weighting. Although these techniques can improve the performance of the ResNet-50 baseline, the effectiveness is limited compared with our strategies.

Our top-1 accuracy is 95.86\%, which obtains a large margin of 10.99\% higher than the second-best methods Triplet loss.
The comparison experiments show that our training strategies facilitate the model to learn a better visual representation, which leads to better generalization and robustness.

\begin{table}[t]
  \centering
  \caption{\wwj{Ablation studies of our semi-supervised classification framework.}}
  \label{table:ablation}
    \begin{tabular}{l|c}
        \toprule
        Method    				& Top-1 (\%) \\
        \midrule
        ResNet-50 (baseline) 	&  79.28 $\pm$ 1.76\\
        \midrule
        MoCo V2 pretraining	& 81.56 $\pm$ 1.36 \\
        MoCo V2 pretraining + Strong Aug. & 76.63 $\pm$ 3.06 \\
        MoCo V2 pretraining + Eq.(\ref{eq:contrastive}) & 75.66 $\pm$ 1.13 \\
        \begin{minipage}{6.5cm} \vspace{0.5mm} MoCo V2 pretraining + $\mathcal{L}^s_{cls}$ \vspace{0.5mm} \end{minipage} & 93.82 $\pm$ 1.57 \\
        \begin{minipage}{6.5cm} \vspace{0.5mm} MoCo V2 pretraining + $\mathcal{L}^u_{cls}$ \vspace{0.5mm} \end{minipage} & 78.65 $\pm$ 1.53 \\
        \begin{minipage}{6.5cm} \vspace{0.5mm} \rv{$\mathcal{L}^s_{cls}$ + $\mathcal{L}^u_{cls}$} \vspace{0.5mm} \end{minipage} & \rv{87.20 $\pm$ 0.93} \\
        \midrule
        \begin{minipage}{6.5cm} \vspace{0.5mm} Final (MoCo V2 pretraining + $\mathcal{L}^s_{cls}$ + $\mathcal{L}^u_{cls}$) \vspace{0.5mm} \end{minipage} & \textbf{95.86} $\pm$ 1.63 \\
        \bottomrule
    \end{tabular}
\end{table}


\begin{figure}[t]
    \centering
    \includegraphics[width=0.99\linewidth]{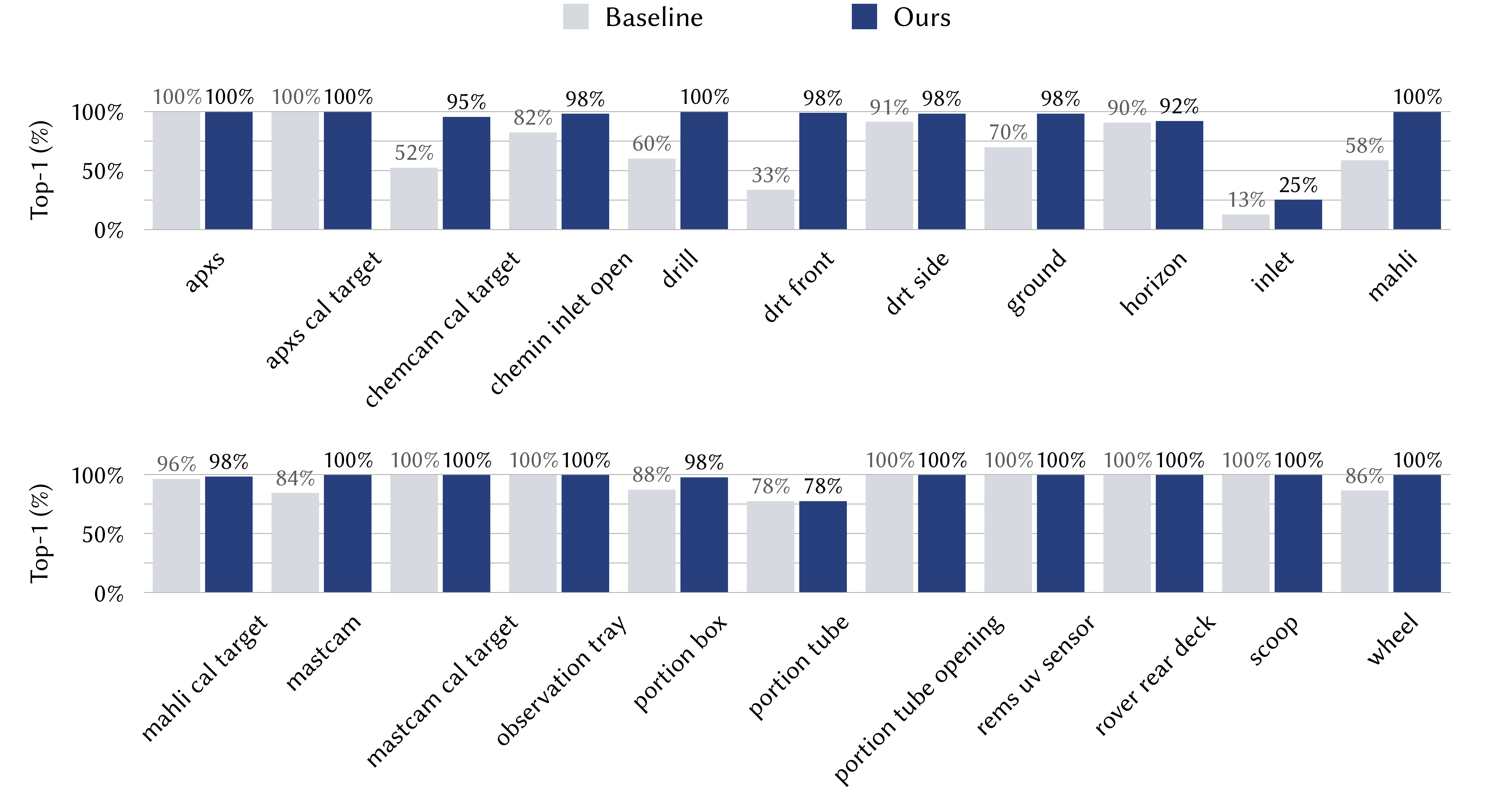}
    \caption{\wwj{Accuracy for each category of the baseline and our MSL rover image classifier.}}
    \label{fig:acc_by_category}
\end{figure}

\subsection{Ablation Studies}

\noindent\textbf{Effect of Learning Strategies.} 
\fzj{
The ablation settings of our designs are shown in Table~\ref{table:ablation}.
%
The accuracy firstly increases by 2.28\% owing to the good feature learned by MoCo V2~\cite{MoCoV2} pretraining.}
The data augmentation we use in contrastive learning is stronger than normal data augmentations used for classification.
If we directly apply our strong augmentation to classification, the performance degrades by 4.93\%, which is because the supervision of classification is too weak to learn against strong augmentation.

\fzj{
Then we study the remarkable effect produced by the supervised inter-class contrastive learning $\mathcal{L}^s_{cls}$. 
The accuracy declines with the original contrastive learning formula, \ie, Eq.(\ref{eq:contrastive}), confirming our motivation of introducing inter-class supervision.
The accuracy also drops with unsupervised similarity learning $\mathcal{L}^u_{cls}$ alone, which is in line with our analysis in Section~\ref{sec:classification}.
\rv{Also, removing MoCo V2 pretraining degrades performance even with $\mathcal{L}^s_{cls}$ and $\mathcal{L}^u_{cls}$.}
The above experiments demonstrate the effectiveness of each component in our semi-supervised learning framework, and all components work together to provide the best performance 95.86\%.}

\wwj{A comparison of the detailed accuracy for each category is shown in Fig.~\ref{fig:acc_by_category}.
Among all 22 categories in the testing set, our model is completely correct in 11 categories.
Compared with the baseline, our model improves the performance by a large margin, especially for the \textit{drt front}, \textit{chemcam cal target}, \textit{mahli}, and \textit{drill classes}.}

\begin{figure}[t]
    \centering
    \includegraphics[width=0.99\linewidth]{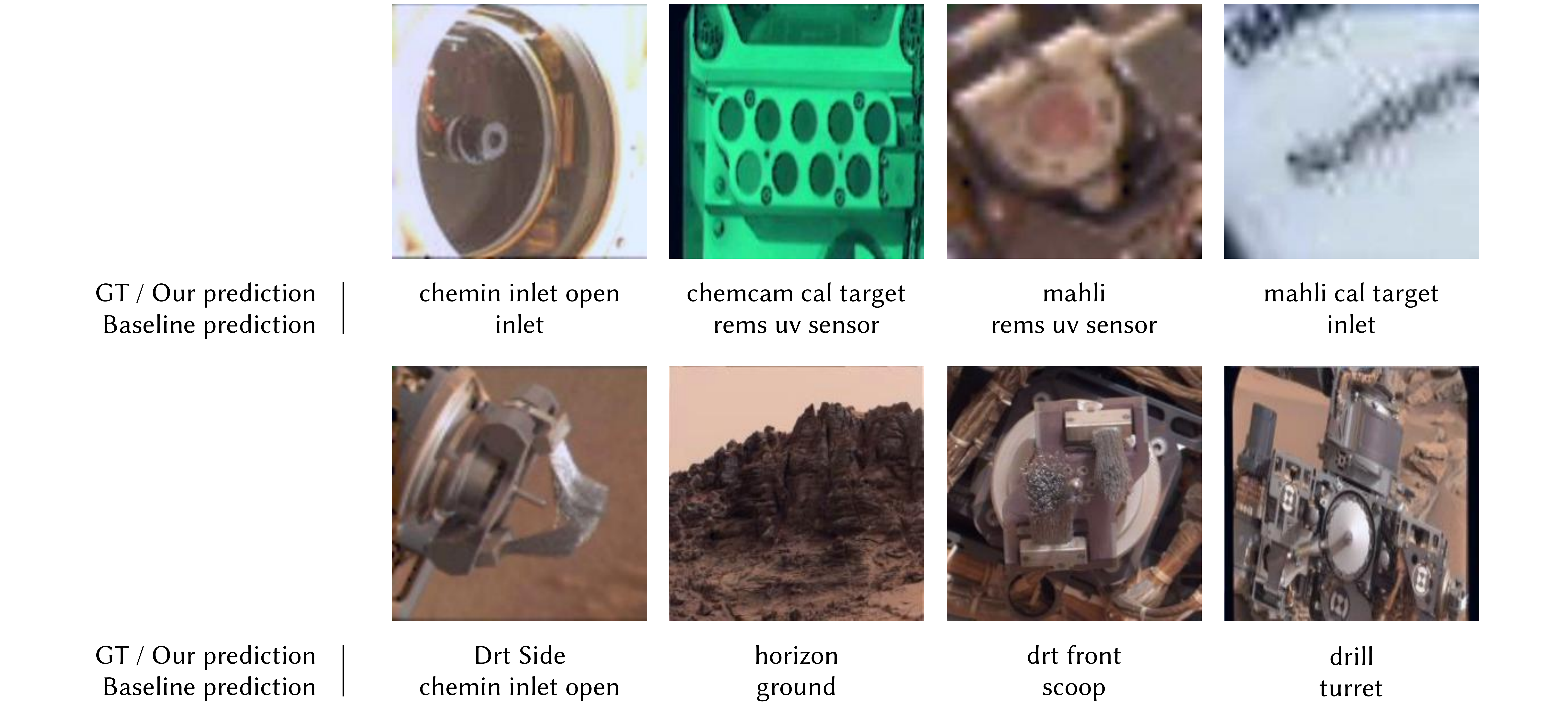}
    \caption{\wwj{Example classification results compared with the baseline.}}
    \label{fig:correct_case_cls}
\end{figure}

\wwj{Some examples can be found in Fig.~\ref{fig:correct_case_cls}.
In the first row, we can see that our classifier is more robust to image quality degradations such as over-exposure, color channel error, and low resolution.
There is a severe train-test object appearance gap in the \textit{drt side} category as stated in Section~\ref{sec:dataset}.
With the proposed semi-supervised learning scheme, this gap can be narrowed as shown in the first column of the second row in Fig.~\ref{fig:correct_case_cls}.
The other images in the second row show that our classifier is more robust to complex devices and terrains.}

\begin{figure}[t]
    \centering
    \includegraphics[width=0.99\linewidth]{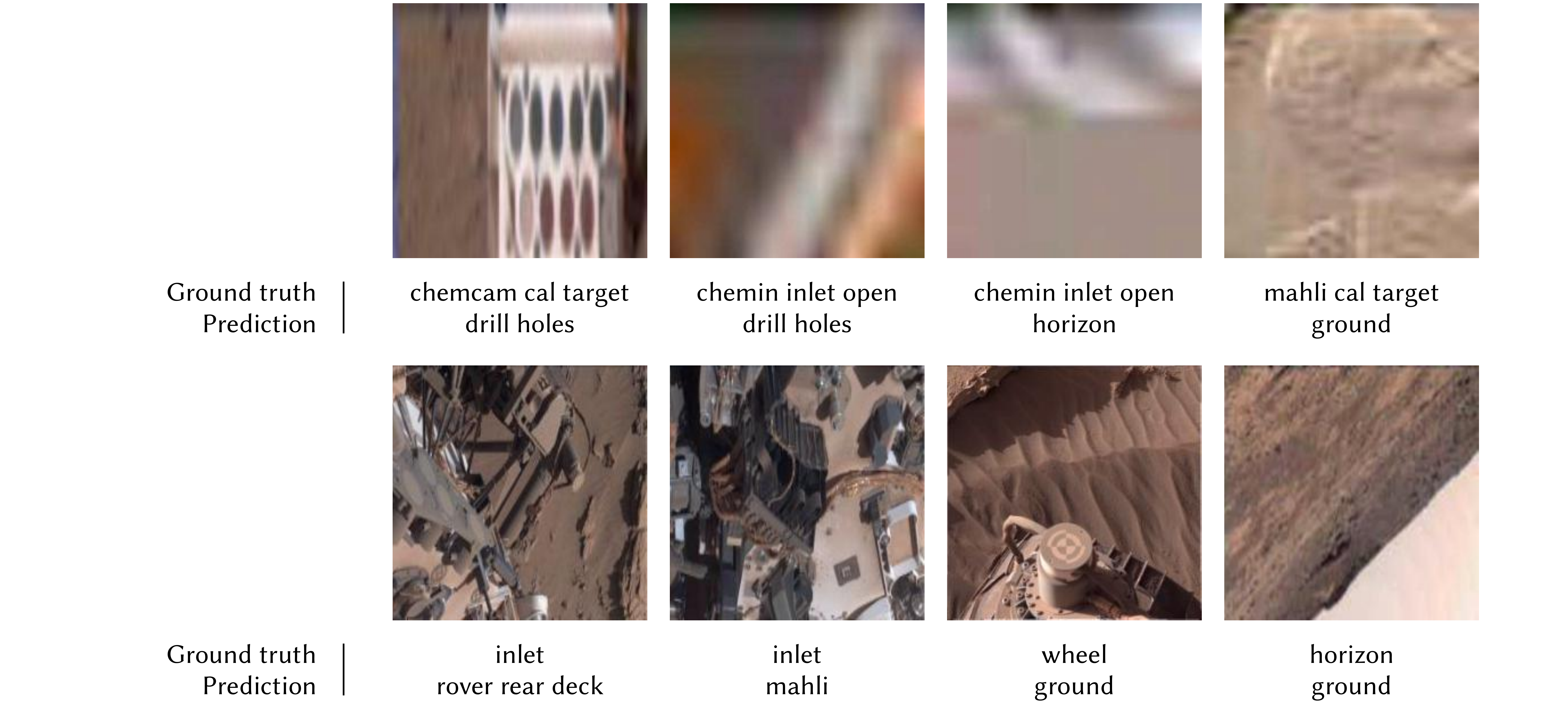}
    \caption{\wwj{Failure cases on MSL rover image classification.}}
    \label{fig:failure_case_cls}
\end{figure}

\begin{table}[t]
\centering
\caption{\wwj{The effect of different classification backbones with our semi-supervised learning. The performance is measured by top-1 (\%) accuracy on the MSL dataset. \rv{All models are based on ImageNet pretraining.}}}
\label{table:backbone_cls}
\begin{tabular}{l|cccc}
    \toprule
       		    & ResNet-50 & VGG-16 & MobileNet V2 & RegNetX-1.6GF \\
    \midrule
    Baseline    &  79.28 $\pm$ 1.76 & 61.99 $\pm$ 2.86 & 80.00 $\pm$ 1.84 & 77.16 $\pm$ 6.65 \\
    Ours        &  \rv{87.20 $\pm$ 0.93} & 65.49 $\pm$ 1.68 &  85.70 $\pm$ 2.22 & 82.94 $\pm$ 1.50 \\
    \midrule
       		    & \rv{ResNet-18} & \rv{ResNet-34}& \rv{ResNet-101} & \rv{ResNet-152} \\
    \midrule
    Baseline    &  \rv{72.69 $\pm$ 1.74} & \rv{78.70 $\pm$ 2.06}  & \rv{81.02 $\pm$ 2.43} & \rv{81.43 $\pm$ 2.29} \\
    Ours        &  \rv{75.43 $\pm$ 1.01} & \rv{80.61 $\pm$ 0.90}  & \rv{87.36 $\pm$ 0.94} & \rv{86.97 $\pm$ 0.67} \\
    \bottomrule
\end{tabular}
\end{table}

\vspace{1em}

\noindent\textbf{Effect of Backbones.} 
\wwj{To evaluate the generalization of our framework, we test other classification backbones besides ResNet-50~\cite{resnet}: VGG-16~\cite{vgg}, MobileNet V2~\cite{MobileNetV2}, \rv{RegNetX-1.6GF~\cite{RegNet}, ResNet-18~\cite{resnet}, ResNet-34~\cite{resnet}, ResNet-101~\cite{resnet}, and ResNet-152~\cite{resnet}} in Table~\ref{table:backbone_cls}.
The baselines are first pretrained on ImageNet then fine-tuned on the MSL dataset.
In the implementation of these frameworks, we do not use the MoCo V2~\cite{MoCoV2} strategy to pretrain for convenience.}

Our framework can improve the classification performance of all backbones, demonstrating the effectiveness of semi-supervised learning.
It also works for lightweight models MobileNet V2 and RegNetX-1.6GF, which may be helpful for designing rover-edge deep models.
\rv{The accuracy of ResNet-50 is better than VGG-16, ResNet-18, and ResNet-34, which is in line with that ResNet-50 is more powerful than them on ImageNet.
The performance is comparable with ResNet-101 and ResNet-152, which might be that ResNet-50 is powerful enough for the Martian classification.}
Two lightweight models perform worse than ResNet-50, indicating that lightweight models are less robust when transferring from ImageNet to Mars data.

\noindent\textbf{Parameter Selection.} \rv{The effect of hyper-parameters $\lambda^s_{cls}$ and $\lambda^u_{cls}$ are shown in Tables~\ref{table:parameter_cls_s} and \ref{table:parameter_cls_u}. Too small values reduce the effect of contrastive learning, and too big values break the balance between different loss terms. Finally, $\lambda^s_{cls}$ = $1$ and $\lambda^u_{cls}$ = $0.2$ achieves the best performance.}

\begin{table}[t]
  \centering
  \caption{Effect of different $\lambda^s_{cls}$ with $\lambda^u_{cls}$ = $0$.}
  \label{table:parameter_cls_s}
    \begin{tabular}{c|ccccc}
        \toprule
        $\lambda^s_{cls}$  & 0.3 & 0.5 & 1.0 & 2.0 & 5.0 \\
        \midrule
        Top-1 (\%)      & 88.40 $\pm$ 0.69
                        & 92.69 $\pm$ 0.67
                        & \textbf{93.82} $\pm$ 1.57
                        & 93.49 $\pm$ 0.27
                        & 93.44 $\pm$ 0.04 \\
        \bottomrule
    \end{tabular}
\end{table}

\begin{table}[t]
  \centering
  \caption{Effect of different $\lambda^u_{cls}$ with $\lambda^s_{cls}$ = $1$.}
  \label{table:parameter_cls_u}
    \begin{tabular}{c|ccccc}
        \toprule
        $\lambda^u_{cls}$  		             & 0 & 0.1 & 0.2 & 0.5 & 1.0 \\
        \midrule
        Top-1 (\%)    & 93.82 $\pm$ 1.57 
                             & 95.43 $\pm$ 0.69
                             & \textbf{95.86} $\pm$ 1.63
                             & 95.33 $\pm$ 1.01
                             & 94.74 $\pm$ 2.10 \\
        \bottomrule
    \end{tabular}
\end{table}

\subsection{Failure Case Study}

\wj{Although our framework has greatly improved the classification performance, it may still make incorrect predictions.}
\wwj{The \textit{inlet} class suffers from a severe train-test gap.
In the training set, the inlet images are mostly close-up. However, in the testing set, the inlets are only parts of the whole image.
Accordingly, the accuracy of the \textit{inlet} class is particularly low as shown in Fig.~\ref{fig:acc_by_category}.}

\wwj{More failure cases are shown in Fig.~\ref{fig:failure_case_cls}.
In the first row, although our classifier is robust to image quality degradation to some extent, it cannot cope with extremely low image quality.
On the left of the second row in Fig.~\ref{fig:failure_case_cls}, we can see that some images contain so many objects that even humans would have trouble classifying them.
Our classifier may also fail when other objects occupy most areas of the image.
As shown on the right of the second row in Fig.~\ref{fig:failure_case_cls}, the ground occupies more areas than the target objects, thus our classifier tends to recognize the images as \textit{ground}.}

\section{Experiments for Mars Imagery Segmentation}
\label{sec:exp_seg}


\wwj{In this section, we evaluate our method under semi-supervised segmentation learning settings on the AI4Mars dataset~\cite{AI4Mars}.}
\wj{We show experimental settings, comparative results, ablation studies, and visual analysis.}

\subsection{Experiment Setup}
\linll{We train the model with \wwj{an} SGD optimizer \wwj{and a} learning rate of 0.01. The polynomial annealing policy is applied for scheduling the learning rate. The batch size is set to 16. We first pretrain the encoder for 60 epochs with \textit{Element-wise Inter-class Contrastive Learning}, \wwj{then} jointly train with contrastive learning loss and segmentation loss for 60 epochs, \wwj{and finally} with \wwj{online pseudo labels} for another 60 epochs.}

\begin{table}[t]
  \centering
  \caption{Segmentation performance on the AI4Mars Dataset. $^\dagger$ indicates ImageNet pretraining.}
  \label{table:seg_comparison}
    \begin{tabular}{l|cc}
        \toprule
        Method & ACC (\%) & mIoU (\%) \\
        \midrule
        Mean Teacher~\cite{antti2017mean} & 72.24 & 55.34\\
        ClassMix~\cite{olsson2021classmix} & 70.67 & 53.84\\
        CutOut~\cite{devries2017improved} & 75.31 & 58.01\\
        CutMix~\cite{CutMix} & 74.81 & 57.70\\
        \midrule
        CCT~\cite{ouali2020semi} & 77.08 & 48.19 \\
        LESS Within-image~\cite{zhao2021contrastive} & 83.65 & 51.24 \\
        LESS Cross-image~\cite{zhao2021contrastive} & 85.37 & 53.60\\
        \midrule
        ReCo~\cite{liu2021bootstrapping} & 72.67 & 55.34 \\
        ReCo~\cite{liu2021bootstrapping} + ClassMix~\cite{olsson2021classmix} & 68.01 & 51.25\\
        ReCo~\cite{liu2021bootstrapping} + CutOut~\cite{devries2017improved} & 75.61 & 58.19\\
        ReCo~\cite{liu2021bootstrapping} + CutMix~\cite{CutMix} & 73.63 & 56.76\\
        \midrule
        ReCo$^\dagger$~\cite{liu2021bootstrapping} & 83.29 & 69.15 \\
        ReCo$^\dagger$~\cite{liu2021bootstrapping} + ClassMix~\cite{olsson2021classmix} & 83.14 & 68.77\\
        ReCo$^\dagger$~\cite{liu2021bootstrapping} + CutOut~\cite{devries2017improved} & 83.23 & 68.73\\
        ReCo$^\dagger$~\cite{liu2021bootstrapping} + CutMix~\cite{CutMix} & 83.11 & 68.78\\
        \midrule
        \textbf{Ours} & \textbf{88.82} & \textbf{70.34}\\
        \bottomrule
    \end{tabular}
\end{table}

\subsection{Comparison Results}
\wwj{We compare our model with state-of-the-art representation learning, augmentation, and contrastive learning methods} \wj{and their combinations.}
\linll{As shown in Table~\ref{table:seg_comparison}, our method achieves the best results on both segmentation accuracy and} \wj{mean Intersection over Union (mIoU).}

\linll{In contrast to Mean Teacher~\cite{antti2017mean}, our method does not directly employ pseudo-labels to train \wj{the segmentation model}.
Instead, pseudo-labels are added to the memory bank for contrastive learning \wwj{and assist \wj{segmentation by extracting} representation through parameter sharing.}
This \wwj{mechanism} can mitigate the negative impact of inaccurate pseudo-labels on \wj{segmentation} performance.}

\linll{The disadvantage of data augmentation methods: CutMix~\cite{CutMix}, CutOut~\cite{devries2017improved}, and ClassMix~\cite{olsson2021classmix}, is that the images of Mars are relatively simple, so the mixed data is not far from other data in the training set. 
\wwj{As a consequence,} the purpose of expanding data distribution cannot be achieved.
Meanwhile, the gap between the training and the testing \wwj{sets} cannot be overcome by \wwj{simply} mixing images, \fzj{but} requires the network to learn better feature representations.
\wwj{In contrast}, our method obtains a better feature space by making full use of the pixels of the training set.
\wwj{Also,} contrastive learning can make the features of different categories more discriminative and \fzj{generalized} to the testing set.}

We also compare semi-supervised segmentation learning methods. CCT~\cite{ouali2020semi} applies unlabeled data to randomly interfere with the features, and constrains its output features to be consistent. However, consistency constraints alone cannot make features more inter-class separable. \rv{LESS~\cite{zhao2021contrastive} applies pixels within and between images for contrastive learning. It directly uses pixel features for building positive and negative samples, which leads to an excessive number of samples and thus makes training unstable. Moreover, the pseudo labels in LESS are fixed once generated. In comparison, we refine the pseudo labels with a retrained encoder for every epoch, which makes the pseudo labels more accurate.
}
\wwj{ReCo~\cite{liu2021bootstrapping} also employs contrastive learning. The consumption of massive computation and memory is the key \fzj{problem} of pixel-level contrastive learning.
To solve this problem, ReCo applies a regional contrast scheme and filters sparse queries and keys based on the confidence of segmentation prediction.
We instead utilize the feature centers of each category as positive and negative samples.
Compared with ReCo, our strategy requires fewer computational resources. Our features are also closer to the clustering center, leading to a better contrastive learning effect.}

\wwj{We also examine the effect of jointly using ReCo and augmentation.
CutOut can slightly improve ReCo, while ClassMix and CutMix may even hurt the performance of ReCo.
ImageNet pretraining has a significant positive effect.
But when pretrained on ImageNet, none of the data augmentation methods can further improve the performance.
Note that our model does not require ImageNet pretraining, but still outperforms methods pretrained on the ImageNet dataset. 
This is because our semi-supervised method can extract more separable features, learning a representation even better than supervised learning on large-scale datasets.}

\begin{figure}[tb]
  \includegraphics[width=\textwidth]{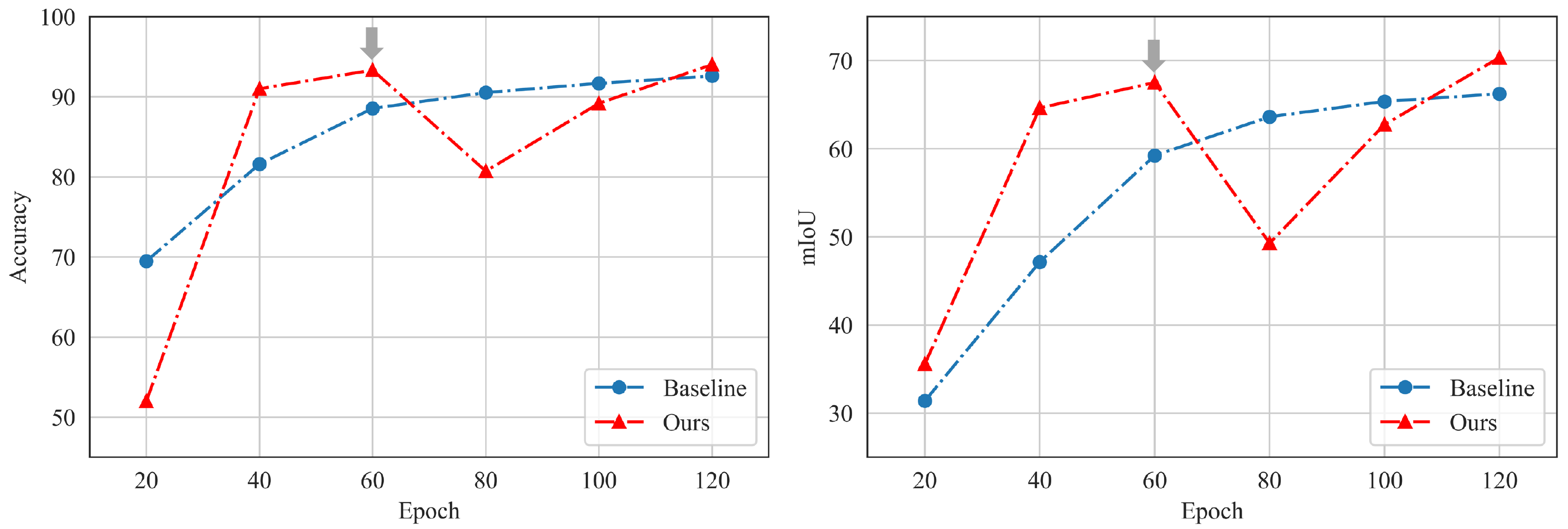}   
  \caption{Curves of segmentation accuracy \wj{(left)} and mIoU \wj{(right)} under different epochs. The gray arrow indicates that we change loss functions at the \wj{60th} epoch.}
  \label{fig:epoch}
\end{figure}

\subsection{Ablation Studies}




\noindent\textbf{Number of Epoch.} \linll{Fig.~\ref{fig:epoch} shows the tendency of accuracy and mIoU of different epochs.}
\wwj{For the first 60 epochs, as the training goes on, there is a significant performance gain.} \linll{With contrastive learning pretraining, the network quickly achieves segmentation accuracy that exceeds the baseline algorithm.}
\wj{Also for the speed of convergence, our network converges at about 60 epochs, while the baseline needs more than 100 epochs.}

\wwj{At the \wj{60th} epoch, we add the semi-supervised loss for co-training.
The performance suddenly drops, which is because the network needs to adapt the feature space to the unlabeled data.}
\linll{Then, as the training progresses}, \wwj{the feature space is gradually refined with the help of the unlabeled data, which narrows the train-test gap and leads to a higher segmentation accuracy and mIoU score.}

\vspace{1em}

\noindent\textbf{Effect of Modules.} \linll{Table~\ref{table:seg_ablation_module} shows the impact of different modules in our design. Accuracy (ACC) is the number of correct pixels divided by the total number of pixels. Mean accuracy (MACC) is the average of the accuracy for each class. The frequency accuracy (FACC) is weighted and summed by the frequency of occurrence of each class. Jointly training the segmentation model with the contrastive loss contributes most to the overall performance gain.}
\wwj{Contrastive learning pretraining improves the mIoU by 1\%, and online pseudo labeling further promotes the comprehensive performance.}


\vspace{1em}

\noindent\textbf{Effect of Threshold.} \wwj{In \textit{Online Pseudo Labeling}, inaccurate labels will bring noise and interfere with the segmentation task.} \linll{Therefore, we employ a threshold to control the pseudo-label assignment of unlabeled data.}
\wwj{Only predictions with confidence higher than the threshold can be appended to the memory bank for contrastive learning.}

\wwj{The effect of this threshold is shown in  Table~\ref{table:seg_ablation_threshold}.}
\linll{We can see that as the threshold decreases, the accuracy and the mIoU of the segmentation task drop along.}
\wwj{Low threshold leads to inaccurate pseudo-label annotations, which become training noises that interfere with segmentation learning.}
\wj{But if the threshold is too high, \eg, 0.99, there will be too few pseudo labels, which provides limited supervision. We finally set the threshold to 0.9.} 



\begin{table}[t]
  \centering
  \caption{Ablation studies of our designs.}
  \label{table:seg_ablation_module}
    \begin{tabular}{l|cccc}
        \toprule
        Method    				& ACC (\%) & MACC (\%) & FACC (\%) & mIoU (\%) \\
        \midrule
        Baseline 	&  92.61 & 71.71 & 86.28 & 66.23\\ 
        \midrule
        \textit{w/o Joint Semi-Supervised Learning} & 92.81 & 72.41 & 86.71 & 66.64\\
        \textit{w/o Pretraining} & 93.78 & 74.75 & 88.39 & 69.30\\
        \textit{w/o Online Pseudo Labeling} & 94.00 & \textbf{75.70} & 88.77 & 70.31\\
        \midrule
        Final Method & \textbf{94.02} & 75.68 & \textbf{88.82} & \textbf{70.34}\\
        \bottomrule
    \end{tabular}
\end{table}

\begin{table}[t]
  \centering
  \caption{The effect of different threshold values.}
  \label{table:seg_ablation_threshold}
    \begin{tabular}{c|cccc}
        \toprule
        Threshold & ACC (\%) & MACC (\%) & FACC (\%) & mIoU (\%) \\
        \midrule
        0.99 & \textbf{94.05} & 75.35 & \textbf{88.85} & 70.14\\
        0.9 & 94.02 & \textbf{75.68} & 88.82 & \textbf{70.34}\\
        0.7 & 94.00 & 75.35 & 88.77 & 70.09\\
        0.5 & 92.89 & 72.61 & 86.81 & 66.92\\
        \bottomrule
    \end{tabular}
\end{table}

\begin{table}[t]
  \centering
  \caption{The effect of different loss weight settings.}
  \label{table:seg_ablation_weight}
    \begin{tabular}{c|cccc}
        \toprule
        $\lambda^s_{seg}$ & ACC (\%) & MACC (\%) & FACC (\%) & mIoU (\%) \\
        \midrule
        
        0.1 & 93.92 & 75.83 & 88.65 & 70.29\\
        0.01 & 93.39 & 75.64 & 88.75 & 70.26\\
        0.001 & 94.02 & \textbf{75.68} & 88.82 & \textbf{70.34}\\
        0.0001 & \textbf{94.04} & 75.46 & \textbf{88.84} & 70.19\\
        \bottomrule
    \end{tabular}
\end{table}

\vspace{1em}

\noindent\textbf{Effect of Loss Weight.} \linll{When jointly training the contrastive learning and segmentation task, the \wwj{balance} between the two-loss functions \wwj{affects} the performance as shown in Table~\ref{table:seg_ablation_weight}.
As the weight of contrastive learning \wwj{$\lambda^s_{seg}$} decreases, the segmentation accuracy becomes higher.
When \wwj{the value of $\lambda^s_{seg}$ is too high}, the network pays more attention to the features required for contrastive learning, while reducing the learning of the segmentation task.}
\wj{But when the value of $\lambda^s_{seg}$ is too low, contrastive learning will have a limited impact on the training process. Finally, we find that $\lambda^s_{seg}=0.001$ achieves the best comprehensive performance.}


\begin{figure}[tb]
  \includegraphics[width=0.99\linewidth]{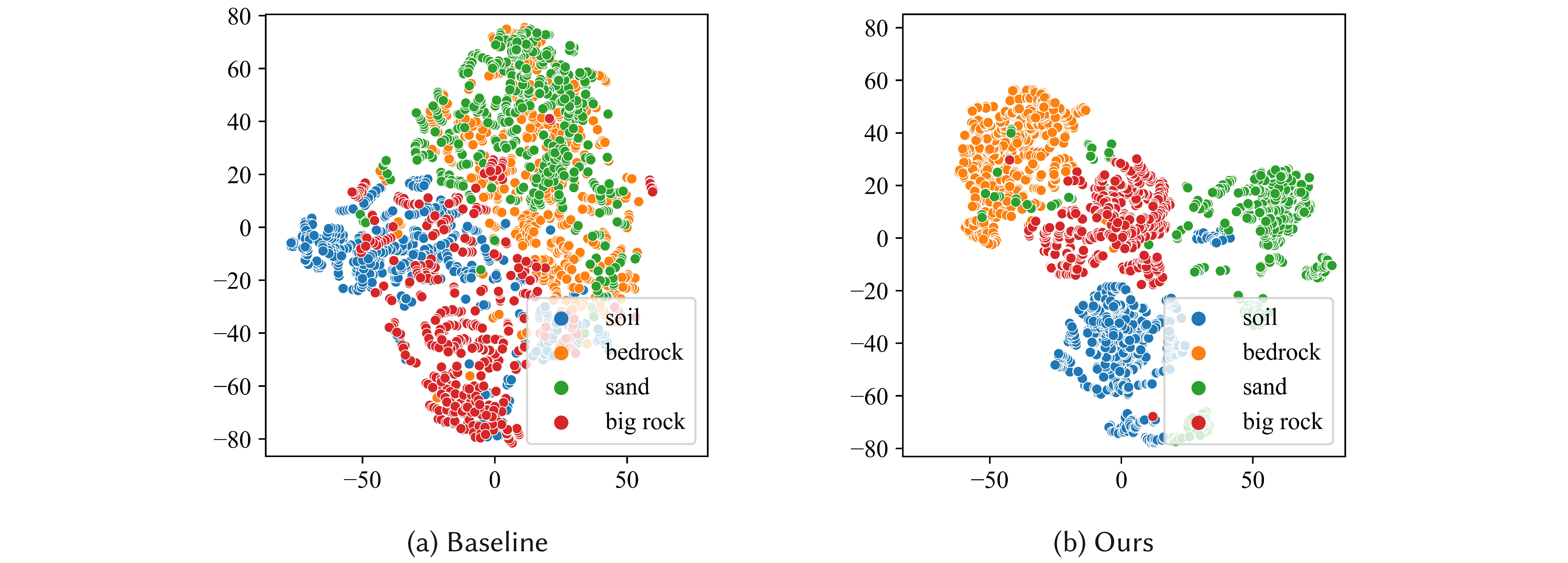}
  \caption{Visualization of the extracted features for each pixel on the AI4Mars dataset.}
  \label{fig:feature}
\end{figure}

\subsection{Visualizations and Interpretability}


\noindent\textbf{Feature Visualization.} \wwj{We visualize the output of the last layer in the encoder $B_{seg}$.}
\linll{As shown in Fig.~\ref{fig:feature}, features extracted by our method are more compact and separable than those extracted by the baseline algorithm. Benefiting from the assistance of contrastive learning, the distance between features of different categories increases, and the features of the same category are more concentrated.}




\wwj{From Fig.~\ref{fig:feature}, we also notice that the feature distance between soil and sand is relatively close, which is in line with that soil and sand have a similar appearance.
Meanwhile, big rocks have diverse appearances.
Accordingly, their representations are more scattered and easier to be confused with other categories.}


\vspace{1em}

\begin{figure}
    \centering
    \includegraphics[width=0.99\linewidth]{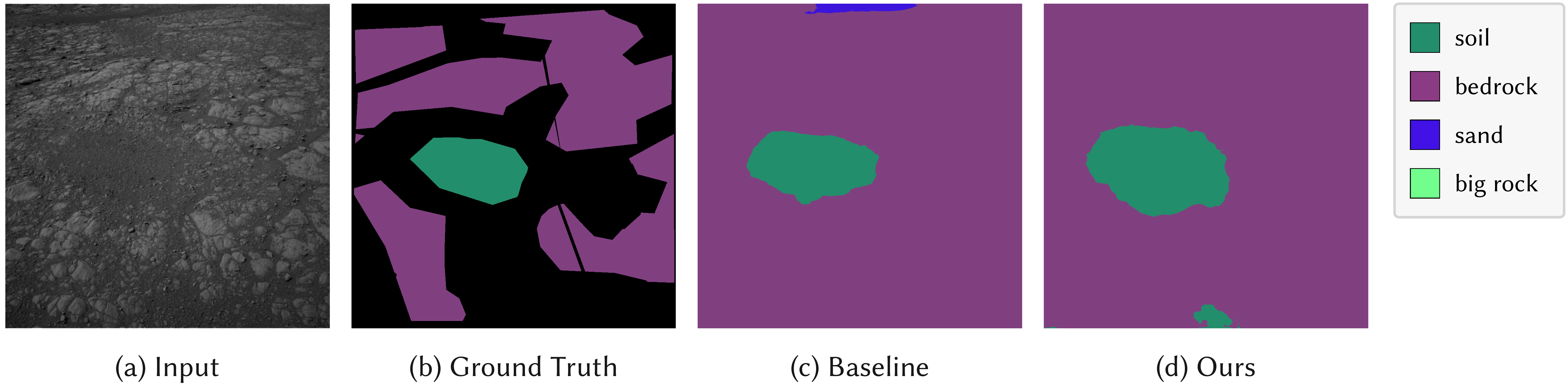}
    \caption{Subjective segmentation results compared with the baseline.}
    \label{fig:subjective_segmentation}
\end{figure}

\noindent\textbf{Subjective Segmentation Results.} 
\wj{We show subjective segmentation results for a testing sample with many unlabeled areas in Fig.~\ref{fig:subjective_segmentation}. Compared with the baseline, our method has more reasonable predictions, such as recognizing the soil at the bottom of the image. In comparison, the baseline may misclassify rocky ground to be sand. The segmentation boundary is also more fine-grained in our prediction.}

\section{Conclusion}
\label{sec:conclusion}

In this paper, we propose a semi-supervised learning framework for Martian machine vision tasks.
For classification, we extend contrastive learning to supervised inter-class and unsupervised similarity-only versions.
For segmentation, we design element-wise contrastive learning and introduce extra supervision by online pseudo labeling.
Experimental results demonstrate the superiority of our designs.
In the future, we will extend our framework to more Martian vision tasks, such as object detection, tracking, and locating.


\bibliographystyle{ACM-Reference-Format}
\bibliography{sample-base}










\end{document}